\pdfoutput=1
\documentclass[preprint,times,12pt]{elsarticle}

\usepackage{amssymb}
\usepackage{times}
\usepackage{subfigure}
\usepackage{soul}
\usepackage{url}
\usepackage[draft]{hyperref}
\usepackage[utf8]{inputenc}
\usepackage[small]{caption}
\usepackage{graphicx}
\usepackage{amsmath,bm}
\usepackage{amsfonts}
\usepackage{booktabs}
\usepackage{natbib}
\usepackage{algorithm}
\usepackage{algpseudocode}
\usepackage{multirow}
\usepackage{pifont}
\newcommand{\cmark}{\ding{51}}%
\newcommand{\xmark}{\ding{53}}%
\usepackage{xcolor}

\usepackage{todonotes}

\usepackage{xcolor}

\usepackage{url}
\usepackage{subfigure}

\makeatletter
\def\ps@pprintTitle{%
 \let\@oddhead\@empty
 \let\@evenhead\@empty
 \def\@oddfoot{}%
 \let\@evenfoot\@oddfoot}
\makeatother

\begin{document}

\begin{frontmatter}



\title{Synthetic-Neuroscore: Using A Neuro-AI Interface for Evaluating\\ Generative Adversarial Networks}


\author[1]{Zhengwei Wang}\ead{zhengwei.wang@tcd.ie, Work done in the Insight Centre for Data Analytics, Dublin City University}
\author[2]{Qi She}\ead{qi.she@intel.com}
\author[3]{Alan F. Smeaton}\ead{alan.smeaton@dcu.ie}
\author[3]{Tom\'as E. Ward}\ead{tomas.ward@dcu.ie}
\author[3]{Graham Healy}\ead{graham.healy@dcu.ie}

\address[1]{V-SENSE, School of Computer Science and Statistics, Trinity College Dublin, Dublin 1, Ireland}
\address[2]{Intel Labs, Beijing, China}
\address[3]{Insight Centre for Data Analytics, Dublin City University, Dublin 9, Ireland}

\begin{abstract}
Generative adversarial networks (GANs) are increasingly attracting attention in the computer vision, natural language processing, speech synthesis and similar domains. Arguably the most striking results have been in the area of image synthesis. However, evaluating the performance of GANs is still an open and challenging problem. Existing evaluation metrics primarily measure the dissimilarity between real and generated images using automated statistical methods. They often require large sample sizes for evaluation and do not directly reflect human perception of image quality. 

In this work, we describe an evaluation metric we call \textbf{Neuroscore}, for evaluating the performance of GANs, that more directly reflects psychoperceptual image quality through the utilization of brain signals. Our results show that Neuroscore has superior performance to the current evaluation metrics in that: (1) It is more consistent with human judgment; (2) The evaluation process needs much smaller numbers of samples; and (3) It is able to rank the quality of images on a per GAN basis. 

A convolutional neural network (CNN) based \textbf{neuro-AI interface} is proposed to predict Neuroscore from GAN-generated images directly without the need for neural responses. Importantly, we show that including neural responses during the training phase of the network can significantly improve the prediction capability of the proposed model. Materials related to this work are provided at \textit{\url{https://github.com/villawang/Neuro-AI-Interface}.}
\end{abstract}




\end{frontmatter}


\section{Introduction}
There is a growing interest in studying generative adversarial networks (GANs) in the deep learning community~\citep{goodfellow2014generative,wang2019generative}. Specifically, GANs have been widely applied to various domains such as computer vision~\citep{karras2018style}, natural language processing~\citep{fedus2018maskgan}, speech synthesis~\citep{donahue2018synthesizing} and time series synthesis~\citep{brophy2019quick}. Compared with other deep generative models (e.g. variational autoencoders (VAEs)), GANs are favored for effectively handling sharp estimated density functions, efficiently generating desired samples and eliminating deterministic bias. Due to these properties GANs have successfully contributed to plausible image generation~\citep{karras2018style}, image to image translation~\citep{zhu2017unpaired}, image super-resolution~\citep{ledig2017photo}, image completion~\citep{yu2018generative} etc. 

However, three main challenges currently in research into GANs could be: (1) Mode collapse -- the model cannot learn the distribution of the full dataset well, which leads to poor generalization ability; (2) Difficult to train -- it is non-trivial for the discriminator and generator in a GAN to achieve Nash equilibrium~\citep{heusel2017gans} during training; (3) Hard to evaluate -- the evaluation of GANs can be considered as an effort to measure the dissimilarity between the real distribution $p_{r}$ and the generated distribution $p_{g}$. Unfortunately, the accurate estimation of $p_{r}$ is intractable. Thus, it is challenging to have a good estimation of the correspondence between $p_{r}$ and $p_{g}$. Challenges (1) and (2) are more concerned with computational aspects where much research has been carried out to mitigate these issues~\citep{li2015generative,salimans2016improved,arjovsky2017wasserstein}. Challenge (3) is similarly fundamental, however limited literature is available and most of the current metrics only focus on measuring the dissimilarity between training and generated images. A more meaningful GAN evaluation metric that is consistent with human perceptions is paramount in helping researchers to further refine and design better GANs.

Although some evaluation metrics, e.g., Inception Score (IS), Kernel Maximum Mean Discrepancy (MMD) and Fr\'echet Inception Distance (FID), have already been proposed~\citep{salimans2016improved,heusel2017gans,borji2018pros}, they have a number of limitations. Firstly, these metrics do not agree with human perceptual judgments and human rankings of GAN models. A small artifact in images can have a large effect on the decision made by a machine learning system~\citep{koh2017understanding}, whilst the intrinsic image content does not change. In this aspect, we consider human perception to be more robust to adversarial images samples when compared to a machine learning system. Secondly, these metrics require large sample sizes for evaluation~\citep{empirical-study,salimans2016improved} and acquiring large-scale samples for evaluation sometimes is not realistic in real-world applications since generating them may be time-consuming.  Finally, the existing metrics are not able to rank individual GAN-generated images by their quality i.e., metrics are generated on a collection of images rather than on a single image basis. The within-GAN variances are crucial because they can provide an insight into the variability of that GAN. 

\begin{figure}[!ht]
		\centering
		\includegraphics[width=\textwidth]{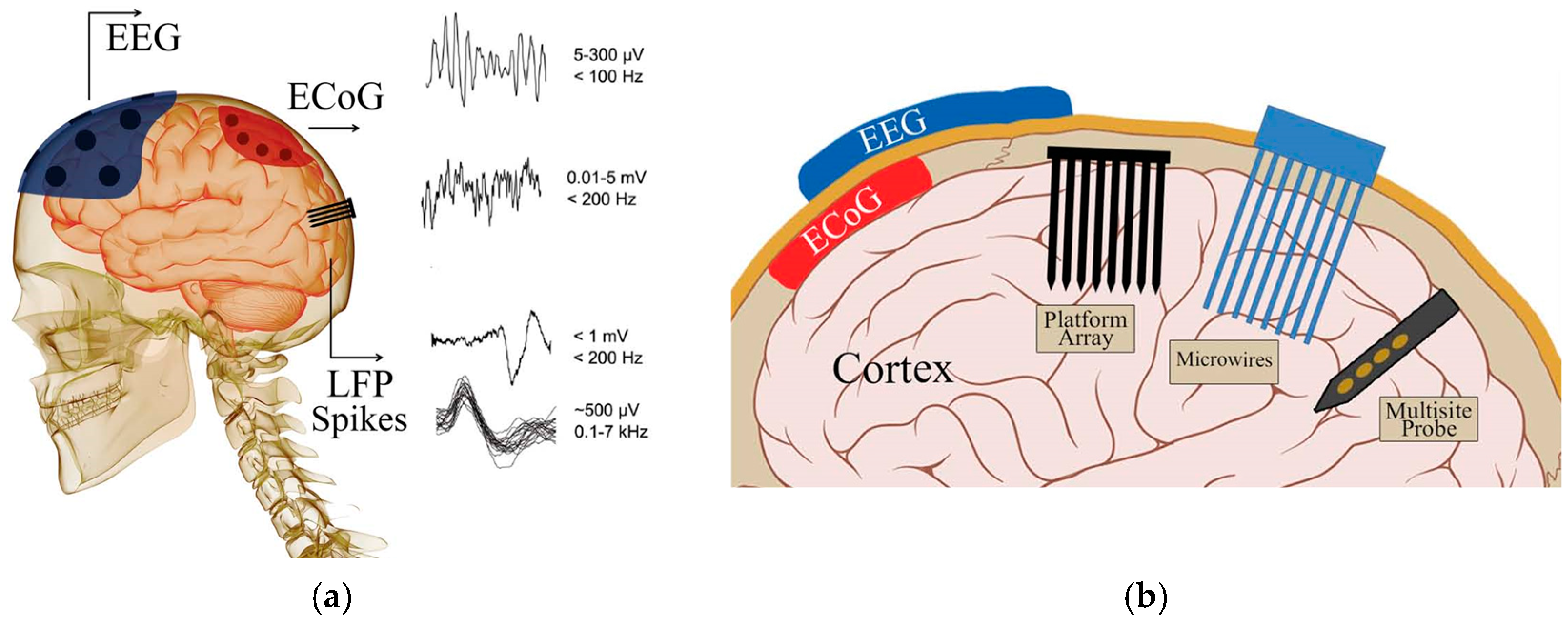}
		\caption{Schematic of different types of recorded neural signals (illustrated in (a)) via invasive and non-invasive measurements (illustrated in (b)). Figure from~\cite{lago2017flexible}.}
		\label{chap07-fig:invasive-noninvasive}
	\end{figure}

The current literature demonstrates that a CNN is able to predict neural responses in the inferior temporal cortex in an image recognition task~\citep{yamins2014performance,yamins2016using} via invasive BCI techniques~\citep{waldert2016invasive}. The ways in which a CNN can be used to predict a neural response with a non-invasive BCI aspect is still an open question. Figure~\ref{chap07-fig:invasive-noninvasive}  
illustrates a schematic of different mesoscopic and macroscopic neural measurement techniques using invasive and non-invasive approaches. In this schematic, only EEG (Electroencephalography) is non-invasively measured from the human scalp~\citep{mouraux2008across}. Other types of neural dynamics such as ECoG and LFP are measured invasively, which requires electrodes to be implanted. Compared to invasively measured neural dynamics, EEG pros are that it is a simple measurement, a non-painful experience during recording, easier to get ethics approval for and more easily generalized to real-world applications. However, EEG suffers challenges such as low signal quality (i.e., low SNR), low spatial resolution (interesting neural activities can span all of the scalp and are thus difficult to localise), all of which makes predicting EEG responses challenging. 

With the success achieved by deep neural networks (DNNs) in areas including computer vision and natural language processing, the operation and functionality of DNNs and its connection with the human brain has been extensively studied and investigated in the literature~\citep{cichy2016comparison,cichy2019deep,groen2018distinct,kuzovkin2018activations,tu2018relating,batista2018deep,yamins2016using,kriegeskorte2015deep,kheradpisheh2016deep}. In this research area, the convolutional neural network (CNN) is widely studied and compared with the visual system in the human brain because both are hierarchical systems and the processing steps are similar. For example in an object recognition task, both CNNs and humans recognize an object by progressively extracting higher-level representations of the visual input through a hierarchy where successive layers operate on the inputs of the proceeding layers e.g., certain patterns of basic shapes, edges and colors as input can be determined at higher levels of the hierarchy to be a particular complex object composed of the inputs. Work reported in~\citep{yamins2016using} outlines a CNN approach to delving even more deeply into understanding the development and organization of sensory cortical processing. It has recently been demonstrated that a CNN is able to reflect the spatio-temporal neural dynamics in the human brain visual processing area~\citep{cichy2016comparison,tu2018relating,kuzovkin2018activations}. Despite much work carried out to reveal the similarity between CNNs and brain systems, research on interactions between CNNs and neural dynamics is limited.

In~\citep{yamins2014performance} the authors demonstrate that a CNN matched with neural data recorded from the inferior temporal cortex of a human subject~\citep{chelazzi1993neural} has high performance in an object recognition task. Given the evidence above that a CNN is able to predict neural responses in the brain, we explore the use of CNNs to predict P300~\citep{polich2007updating,carrillo2000effect} amplitudes in this paper. This type of model can then produce (synthetic) EEG feedback for different types of GANs.

By applying advanced statistical and machine learning techniques to non-invasive EEG, better source localization and reconstruction becomes possible. Our previous work~\citep{wang2018spatial,wang2018review} demonstrated the effectiveness of using spatial filtering approaches for reconstructing P300 source ERP signals. Remaining low SNR issues can be further remedied by averaging EEG trials. Based on this evidence, we explore the use of DNNs to predict a metric we call Neuroscore~\citep{wang2018use}, when neural information is available via EEG. 

\begin{figure}[ht!]
    \centering
    \includegraphics[width=1\textwidth]{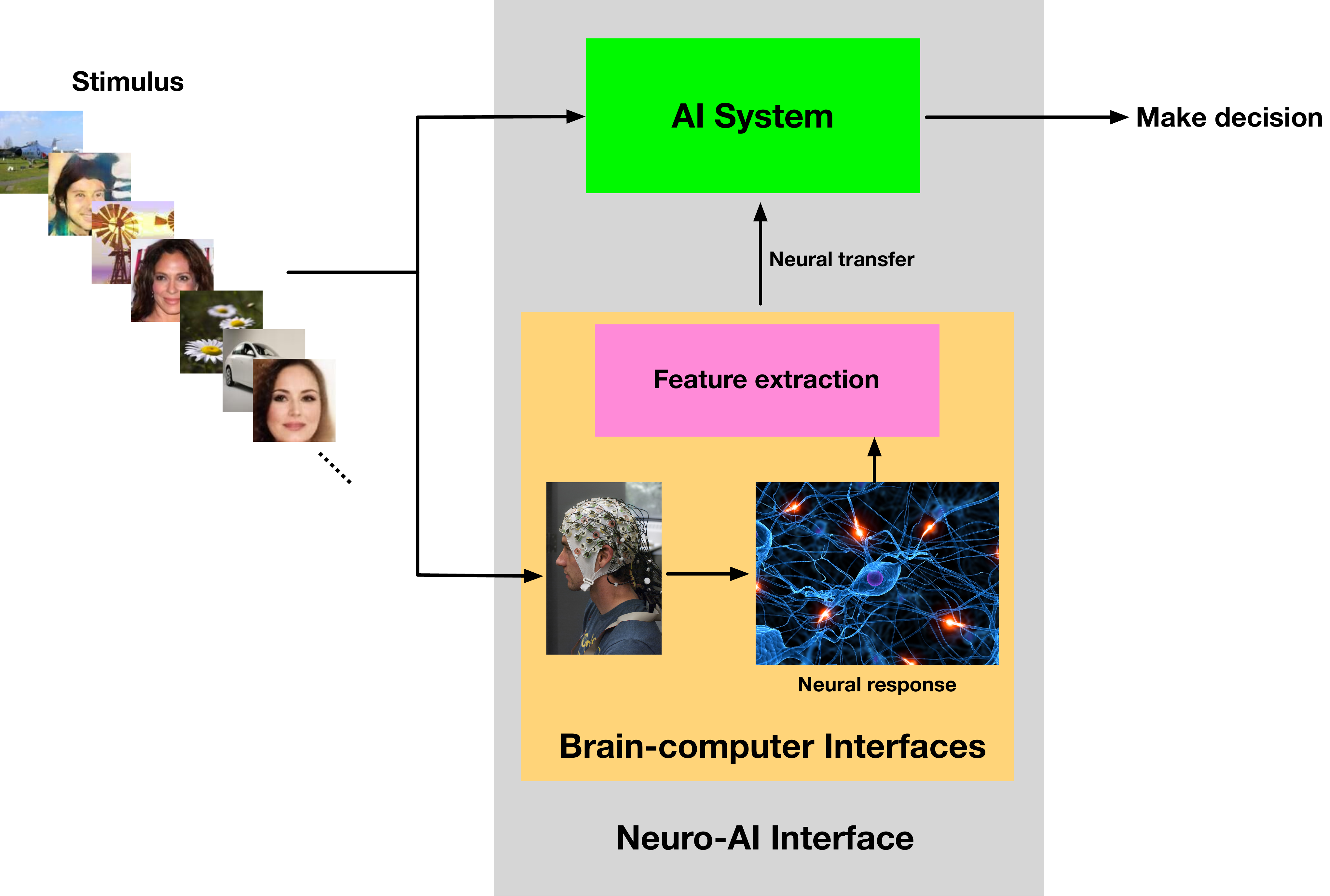}
    \caption{Schematic of a neuro-AI interface. Stimuli (image stimuli used in this work) are simultaneously presented to an AI system and to participants. Participants' neural responses are transferred to the AI system as supervised information for assisting the AI system to make decision.}
    \label{fig:neuron_AI_interface1}
\end{figure}

In this work, we describe a metric called Neuroscore to evaluate the performance of GANs, which is derived from a neurophysiological response recorded via non-invasive electroencephalography (EEG). We demonstrate and validate a neural-AI interface (as seen in Figure~\ref{fig:neuron_AI_interface1}), which uses neural responses as supervisory information to train a CNN. The trained CNN model is then able to predict Neuroscore for images without requiring the corresponding neural responses. We test this framework via three models: Shallow convolutional neural network, Mobilenet V2~\citep{sandler2018mobilenetv2} and Inception V3~\citep{szegedy2016rethinking}. 

In outline, Neuroscore is calculated via measurement of the P300, an event-related potential (ERP) present in EEG, via a rapid serial visual presentation (RSVP) paradigm. The P300 and RSVP paradigm are mature techniques in the brain-computer interface (BCI) community and have been applied in a wide variety of tasks such as image search \citep{gerson2006cortically}, information retrieval \citep{mohedano2015exploring}, and others. The unique benefit of Neuroscore is that it more directly reflects human perceptual judgment of images, which is intuitively more reliable compared to conventional metrics in the literature~\citep{borji2018pros}. In summary, our contributions are two-fold: 
\begin{itemize}
    \item  We combine human perception research with GANs and deep learning research. This represents a new avenue of investigation in the development of better GANs technologies.
    \item We propose a type of neuro-AI interface and training strategy to generalize the use of Neuroscore, which can be directly used for GAN evaluations without recording EEG. This enables our Neuroscore to be more widely applied to real-world scenarios, with a new measure we name synthetic-Neuroscore.
\end{itemize}

\section{Related Work}
Three well-known metrics are compared with Neuroscore in this paper. 
\subsection{Inception Score (IS)}
Inception Score is the most widely used metric in the literature \citep{salimans2016improved,empirical-study,borji2018pros}. It uses a pre-trained Inception network \citep{szegedy2016rethinking} as an image classification model $\mathcal{M}$ to compute
\begin{equation*} \label{eq:IS-formula}
	\mathrm{IS} = \exp\left(\mathbb{E}_{\bm{\mathrm{x}}\sim p_{g}} \left[\mathrm{KL}\left(p_{\mathcal{M}}\left(\mathrm{y} \vert \bm{\mathrm{x}}\right)\vert \vert p_{\mathcal{M}}(\mathrm{y})\right)\right]\right)
\end{equation*}
where $p_{\mathcal{M}}(\mathrm{y} \vert \bm{\mathrm{x}})$ is the label distribution of $\bm{\mathrm{x}}$ that is predicted by the model $\mathcal{M}$ and $p_{\mathcal{M}}(\mathrm{y})$ is the marginal probability of $p_{\mathcal{M}}(\mathrm{y} \vert \bm{\mathrm{x}})$ over the probability $p_{g}$. A larger inception score will have $p_{\mathcal{M}}(\mathrm{y} \vert \bm{\mathrm{x}})$ close to a point mass and $p_{\mathcal{M}}(\mathrm{y})$ close to uniform, which indicates that the Inception network is very confident that the image belongs to a particular ImageNet category~\citep{deng2009imagenet} where all categories are equally represented. This suggests the generative model has both high quality and diversity. 

\subsection{Kernel Maximum Mean Discrepancy (MMD)}
MMD is a method for comparing two distributions, in which the test statistic is the largest difference in expectations over functions in the unit ball of a reproducing kernel Hilbert space~\citep{gretton2012kernel}. MMD is computed as 
\begin{equation*} \label{eq:MMD-formula}
\begin{aligned}
	\mathrm{MMD}^{2}(p_{r}, p_{g})  =  \mathbb{E}_{ {\substack{\bm{\mathrm{x}}_{r}, \bm{\mathrm{x}}_{r}^\top  \sim p_{r}  \\ {\bm{\mathrm{x}}_{g}, \bm{\mathrm{x}}_{g}^\top \sim p_{g} } } } } {\left[k(\bm{\mathrm{x}}_{r}, \bm{\mathrm{x}}_{r}^\top) - 2k(\bm{\mathrm{x}}_{r}, \bm{\mathrm{x}}_{g}) + k(\bm{\mathrm{x}}_{g}, \bm{\mathrm{x}}_{g}^\top)\right]}
\end{aligned}
\end{equation*}
It measures the dissimilarity between $p_{r}$ and $p_{g}$ for some fixed kernel function $k$, such as a Gaussian kernel~\citep{li2015generative}. A lower MMD indicates that $p_{g}$ is closer to $p_{r}$, showing the GAN has better performance.

\subsection{Fr\'echet Inception Distance (FID)} 
FID uses a feature space extracted from a set of generated image samples by a specific layer of the Inception network~\citep{heusel2017gans}. The feature space is modelled via a multivariate Gaussian by the mean $\bm{\mathrm{\mu}}$ and covariance $\bm{\mathrm{\Sigma}}$. FID is computed as
\begin{equation*} \label{eq:FID-formula}
	\mathrm{FID}(p_{r}, p_{g}) = \vert \vert \bm{\mathrm{\mu}}_{r} - \bm{\mathrm{\mu}}_{g} \vert \vert_{2}^{2} + \mathrm{Tr}\left(\bm{\mathrm{\Sigma}}_{r} + \bm{\mathrm{\Sigma}}_{g} - 2(\bm{\mathrm{\Sigma}}_{r}\bm{\mathrm{\Sigma}}_{g}\right)^{\frac{1}{2}})
\end{equation*}
Similar to MMD, lower FID is better, corresponding to more similar real and generated samples as measured by the distance between their activation distributions.

For Inception Score, the score is calculated through the Inception model~\citep{szegedy2016rethinking}. It has been shown that Inception Score is very sensitive to the model parameters \citep{barratt2018note}. Even the score produced by the same model trained using different libraries (e.g., Tensorflow, Keras, PyTorch) differ a lot from each other. It also requires a large sample size for the accurate estimation for $p_{\mathcal{M}}(\mathrm{y})$. FID and MMD both measure the similarity between training images and generated images based on the feature space~\citep{empirical-study}, since the pixel representations of images do not naturally support for meaningful Euclidean distances to be computed ~\citep{forsyth2003modern}. The main concern about these two methods is whether the distributional characteristics of the feature space exactly reflect the distribution for the images~\citep{koh2017understanding}.  

We list the supported features of Neuroscore and traditional metrics in Table~\ref{tab:metric-compare}. Neuroscore can not only evaluate image quality as can the other metrics, but also have $3$ unique characteristics, which will be demonstrated in Section~\ref{results}.
\begin{table}[!htbp]
    \centering
    \begin{tabular}{c|c|c|c|c}
        \hline
        {Feature} & {IS} & {MMD} & {FID} & {\textbf{Neuroscore}}\\ \hline
        {Evaluate image quality} & {\cmark} & {\xmark} & {\cmark} & {\cmark}\\ \hline  
        {Consistent with human} & {\xmark} & {\xmark} & {\xmark} & {\cmark}\\ \hline
        {Small sample size} & {\xmark} & {\xmark} & {\xmark} & {\cmark}\\ \hline
        {Rank images} & {\xmark} & {\xmark} & {\xmark} & {\cmark}\\ \hline
    \end{tabular}
    \caption{Comparison between Neuroscore and other metrics.}
    \label{tab:metric-compare}
\end{table}

\section{Preliminaries}
\subsection{Generative Adversarial Networks}
A generative adversarial network (GAN) has two components, the discriminator $D$ and the generator $G$. Given a distribution $\pmb{z} \sim p_{\pmb{z}}$, $G$ defines a probability distribution $p_{g}$ as the distribution of the samples $G(\pmb{z})$. The objective of a GAN is to learn the generator's distribution $p_{g}$ that approximates the real data distribution $p_{r}$. Optimization of a GAN is performed with respect to a joint loss for $D$ and $G$ as 
\begin{equation*} \label{eq:GAN-formula}
	\min \limits_{G} \max \limits_{D} \mathbb{E}_{\bm{\mathrm{x}} \sim p_{r}} \mathrm{log}[D(\bm{\mathrm{x}})] + \mathbb{E}_{\bm{\mathrm{z}} \sim p_{\bm{\mathrm{z}}}} \mathrm{log}\left[1 - D(G(\bm{\mathrm{z}}))\right]
\end{equation*}

\subsection{P300 (or P3) Component and Preprocessing} \label{sec:P300}
In neuroscience, the P300 ERP component refers to a voltage change measured on the scalp which arises from current flow changes in the brain in response to a target stimulus~\citep{polich2007updating}, that can be measured with EEG. It reflects a participant's attention, which can be modulated by the specific instruction given to a participant. It has been shown in previous studies that real face stimuli generate larger P300/LPP potentials than non-real face stimuli such as cartoon face images \citep{schindler2017differential,ling2012comparative,zhao2019event}. Furthermore, the P300/LPP increases linearly with face realism, reflecting increased activity in visual and parietal cortex for more realistic faces\citep{schindler2017differential}. The P300 response elicited by a target stimulus is typically evident between $300$ -- $600$ ms post stimulus presentation depending on the type of task. EEG is normally recorded by using multiple channels e.g.. 32 channels, which makes it difficult to estimate the P300 source amplitude. We use an LDA beamformer \citep{treder2016lda,wang2018spatial} to reconstruct the P300 source signal from the recorded raw EEG epochs.

Briefly, given a target EEG epoch $\bm{\mathrm{X}}_{i} \in \mathbb{R}^{C \times T}$ and a standard EEG epoch\footnote{A target EEG epoch is an EEG trial (time duration 0 -- 1 s) which corresponds to a target stimulus i.e., DCGAN, BEGAN, PROGAN and RFACE images in this study. A standard/non-target EEG epoch is an EEG trial which corresponds to a non-target images i.e., non-face image in this work.} $\bm{\mathrm{K}}_{i} \in \mathbb{R}^{C \times T}$ ($C$ is the number of channels and $T$ is time points in each EEG epoch). The optimization problem for the LDA beamformer is to find a projection vector $\bm{\mathrm{w}} \in \mathbb{R}^{C \times 1}$ that solves the optimization problem:
\begin{equation} \label{eq:beam_cost_function}
	\min \limits_{\bm{\mathrm{w}}} \bm{\mathrm{w}}^\top \bm{\mathrm{\Sigma}} \bm{\mathrm{w}} \hspace{5pt} \mathrm{s.t.} \bm{\mathrm{w}}^\top\bm{\mathrm{p}}=1
\end{equation}
where $\bm{\mathrm{\Sigma}} \in \mathbb{R}^{C \times C}$ is the EEG epoch covariance matrix ($\bm{\mathrm{\Sigma}} = \frac{1}{N} \sum_{i=1}^{N}\bm{\mathrm{X}}_{i}\bm{\mathrm{X}}_{i}^\top$, $N$ is number of trials) and $\bm{\mathrm{p}}\in \mathbb{R}^{C \times 1}$ is the spatial pattern difference between target and standard condition~\citep{treder2016lda}. The closed-form solution is
\begin{equation}\label{eq:w}
	\bm{\mathrm{w}}=\bm{\mathrm{\Sigma}}^{-1} \bm{\mathrm{p}} ( \bm{\mathrm{p}}^\top \bm{\mathrm{\Sigma}}^{-1} \bm{\mathrm{p}}) ^{-1}
\end{equation}
The source signal of each single-trial $\bm{\mathrm{s}}$ can be obtained as
\begin{equation}\label{eq:S_P300}
	\bm{\mathrm{s}}= \bm{\mathrm{w}}^\top \bm{\mathrm{X}}_{i} =
	(\bm{\mathrm{p}}^\top \bm{\mathrm{\Sigma}}^{-1} \bm{\mathrm{p}} )^{-1} \bm{\mathrm{p}}^\top \bm{\mathrm{\Sigma}}^{-1} \bm{\mathrm{X}}_{i}
\end{equation}
where $\bm{\mathrm{s}} \in \mathbb{R}^{1 \times T}$. Hence, LDA beamformer enables transformation of multi-channel EEG epochs to single-channel EEG epochs facilitating more robust measurement of the P300 and its amplitude. 

\section{Methodology}
\subsection{Data Acquisition and Experiment}
\begin{figure}[ht!]
\centering
 \subfigure{\includegraphics[width=.2\textwidth]{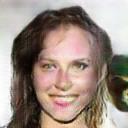} }
 \hspace{5pt}
 \subfigure{\includegraphics[width=.2\textwidth]{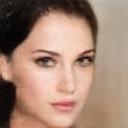} }
 \hspace{5pt}
 \subfigure{\includegraphics[width=.2\textwidth]{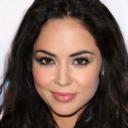} }
 \hspace{5pt}
 \subfigure{\includegraphics[width=.2\textwidth]{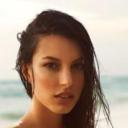} }
\caption{Face image examples used in the experiments. From left to right: DCGAN, BEGAN, PROGAN, and real face (RFACE).} \label{fig-face-example}
\end{figure}

We used three GAN models to generate synthetic images of faces: DCGAN~\citep{radford2015unsupervised}, BEGAN~\citep{berthelot2017began} and progressive growing of GANs (PROGAN)~\citep{karras2017progressive} with sample outputs shown in Figure~\ref{fig-face-example}.
Image streams in the experiment contain generated images from DCGAN, BEGAN and PROGAN, as well as real face (RFACE) images and non-face category images. RFACE images were sampled from the CelebA dataset~\citep{liu2015faceattributes}. Non-face category (standard images) were sampled from the ImageNet dataset~\citep{deng2009imagenet}, similar to those used in other RSVP experiments such as~\citep{healy2017eeg,nails}. EEG data for 12 participants was gathered. Data collection was carried out with approval from Dublin City University Research Ethics Committee (REC/2018/115). Each participant completed two types of task which we call the behavioral experiment (BE) task and the rapid serial visual presentation (RSVP) task. The sequence of blocks presented in the experiment was: BE $\rightarrow$ RSVP $\rightarrow$ BE $\rightarrow$ RSVP $\rightarrow$ BE. The presented images were randomly shuffled (across and within blocks), meaning the appearance of face images could not be predicted ahead of time by a participant i.e., they occurred at random times but always in the same quantity.

The objective of the BE task was to record participants' responses to each type of image category while the RSVP task was to record EEG when participants were viewing the rapid presentation of images. The ultimate goal of this study was to compare whether the EEG responses in the RSVP task were consistent with participants' responses in the BE task.

\begin{figure}[ht!]
	\centering
	\includegraphics[width=.7\textwidth]{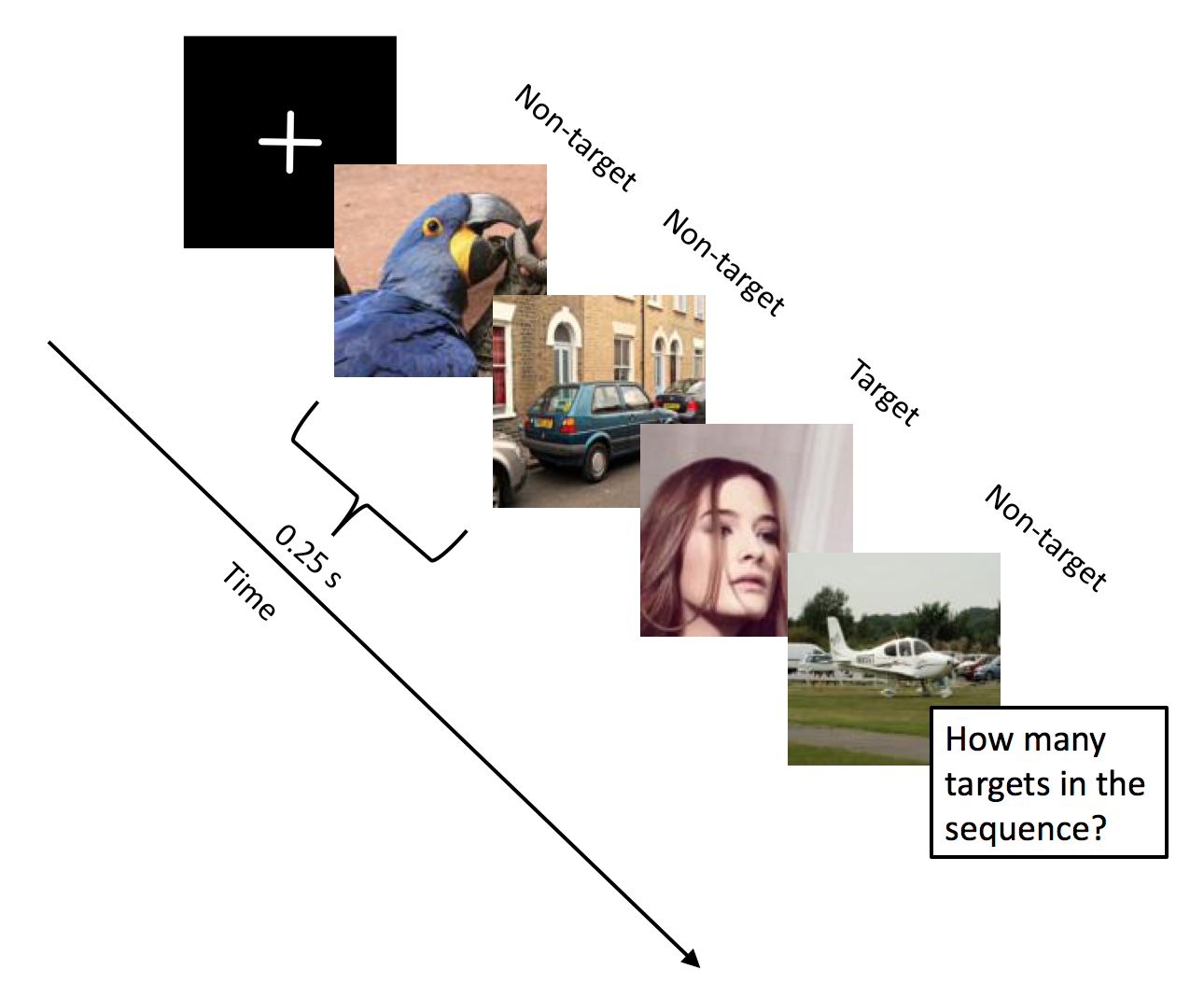}
	\caption{An example of RSVP experimental protocol used in this work. A rapid image stream containing targets and standards (non-target) is presented to participants at 4 Hz (4 images per second) presentation rate.}
	\label{fig:rsvp_paradigm}
\end{figure}

An example of the RSVP experimental protocol is shown in Figure~\ref{fig:rsvp_paradigm}.
The RSVP task contained 26 blocks. Each RSVP block contained 240 images (6 images for each face category thus 24 face targets in total and 216 non-face images), thus there were 6,240 images (624 face targets / 5,616 non-face images) available for each participant. In the RSVP task, image streams were presented to participants at a 4 Hz presentation rate. Participants in RSVP blocks were asked to search for real face (RFACE) images\footnote{P300 responses were elicited for all GAN image categories e.g., while DCGAN had almost perfect behavioral accuracy labelled as being ‘fake’, DCGAN images still elicited a P300.}. This instruction to participants was constructed so that they would maintain attention to detect face images (from all GAN types), and furthermore focus their attention to what they perceived as real face images~\citep{carrillo2000effect}. Details of the experiment can be found in~\citep{wang2018use}.

EEG was recorded from participants in both  the BE and RSVP tasks along with timestamp information for image presentation and behavioural responses (via a photodiode and hardware trigger) to allow for precise epoching of the EEG signals for each trial~\citep{wang2016investigation}. EEG data was acquired using a 32-channel BrainVision actiCHamp at 1,000 Hz sampling frequency, using electrode locations as defined by the International 10-20 system. To enhance the low signal-to-noise ratio of the acquired EEG, pre-processing is required. Pre-processing typically involves re-referencing, filtering the signal (by applying a bandpass filter to remove environmental noise or to remove activity in non-relevant frequencies), epoching (extracting a time epoch typically surrounding the stimulus onset) and trial/channel rejection (to remove those containing artifacts). In this work, a common average reference (CAR) was utilized and a bandpass filter (i.e., 0.5-20 Hz) was applied prior to epoching. EEG data was then downsampled to 250 Hz. Only trials where behavioral responses occurred between 0 and 1 second after the presentation of a stimulus were used. Trial rejection was carried out to remove those trials containing noise such as eye-related artifacts (via a peak-to-peak amplitude threshold across all electrodes).

\subsection{Neuroscore}
We used a rapid serial visual presentation (RSVP) paradigm~\citep{wang2016investigation,wang2018review,healy2020experiences} to elicit the P300 ERP. Our experimental procedure is illustrated in our previous published work~\citep{wang2018use}. We average the single-trial P300 amplitude (as Neuroscore) to mitigate the background EEG noise~\citep{polich2007updating}, which renders a stable measurement of the EEG response to a typical type of stimulus. In general, our Neuroscore is calculated via two steps: (1) Reconstruct the P300 source signal from the raw EEG; (2) Average the P300 amplitude of each reconstructed single-trial source signal across trials (see Algorithm~\ref{al:calculate-NS}).
\begin{algorithm}[!ht]
    \caption{Calculation of Neuroscore}
    \label{al:calculate-NS}
    \begin{algorithmic}[1]
    \item[\textbf{Input:}]
    \begin{itemize}
        \Statex \item $\bm{\mathrm{X}} \in \mathbb{R}^{N \times C \times T}$ is the EEG signal corresponding to the target stimulus, where $N$ is the number of target trials, $C$ is the number of channels, and $T$ is the number of time points.
    	\item $\bm{\mathrm{K}} \in \mathbb{R}^{M \times C \times T}$ is the EEG signal corresponding to the standard stimulus, $M$ is number of standard trials, $C$ is number of channels, $T$ is number of time points. The target and standard EEG trials are already explained in section~\ref{sec:P300}.
    \end{itemize}
    \item[\textbf{Output:}] Neuroscore
    \State $\bm{\mathrm{\Sigma}} = \frac{1}{N} \sum_{i=1}^{N}\bm{\mathrm{X}}_{i}{\bm{\mathrm{X}}_{i}}^\top + \frac{1}{M} \sum_{i=1}^{M}\bm{\mathrm{K}}_{i}{\bm{\mathrm{K}}_{i}}^\top$
    \For{$t_{i}$ in [$400$ ms, $600$ ms]}
        \State $\bm{\mathrm{p}}=\frac{1}{N}\sum_{i=1}^{N}\bm{\mathrm{X}}_{i, t_{i}} - \frac{1}{M} \sum_{i=1}^{M}\bm{\mathrm{K}_{i, t_{i}}}$
        \State {$\bm{\mathrm{w}}=\bm{\mathrm{\Sigma}}^{-1} \bm{\mathrm{p}} ( \bm{\mathrm{p}}^\top \bm{\mathrm{\Sigma}}^{-1} \bm{\mathrm{p}}) ^{-1}$}
        \State {$\mathrm{J}_{t_{i}} \gets \bm{\mathrm{w}}^\top \bm{\mathrm{\Sigma}} \bm{\mathrm{w}}$}
        \State{$\mathrm{W}_{t_{i}} \gets \bm{\mathrm{w}}$}
    \EndFor
    \State $t_{optimal}$=$\mathrm{argmin}_{t_{i}} \mathrm{J}$
    \State $\bm{\mathrm{w}}_{optimal}$=$\mathrm{W}_{t_{optimal}}$
    \State $\mathrm{t}_{P300}$=[$\mathrm{t}_{optimal}$ - $100$ ms, $\mathrm{t}_{optimal}$ + $100$ ms] \Comment \textit{This is time window being detected for P300.} 
    \For{$i = 1:N$}
        \State{$\bm{\mathrm{s}} = \bm{\mathrm{w}}_{optimal}^\top\bm{\mathrm{X}}_{i}$}
        \State{$a = \max(\bm{\mathrm{s}}_{t_{p300}}$)}
        \State{$\mathrm{A}_{i} \gets$ a} 
    \EndFor
    \State $\mathrm{Neuroscore}$ = $\dfrac{1}{N}\sum_{i=1}^{N}\mathrm{A}_{i}$
    \end{algorithmic}
\end{algorithm}

\begin{figure}[ht!]
	\centering
	\subfigure[Reconstructed source P300 signals for each type of image category by using LDA beamformer across 12 participants. P300 component appears in 400 ms -- 600 ms. Solid lines are averaged responses across participants while shadow areas represent the standard deviation. ]{\includegraphics[width=.6\textwidth]{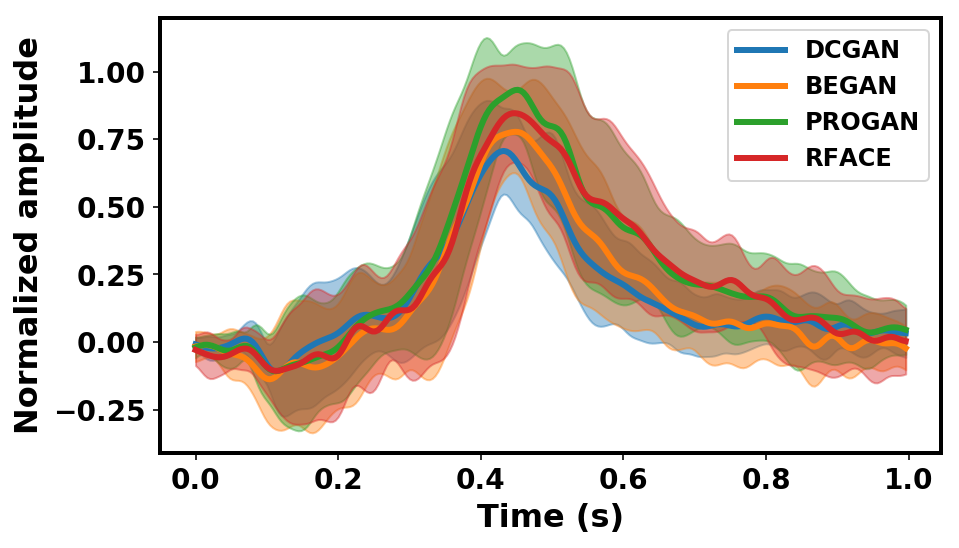}}
	
	\subfigure[Correlation between Neuroscore and human judgment (i.e., participants' behavioural accuracy) across participants. \textbf{Pearson correlation statistics is $r(36) = -0.828, p = 4.766 \times 10^{-10}$}.]
	{\includegraphics[width=.7\textwidth]{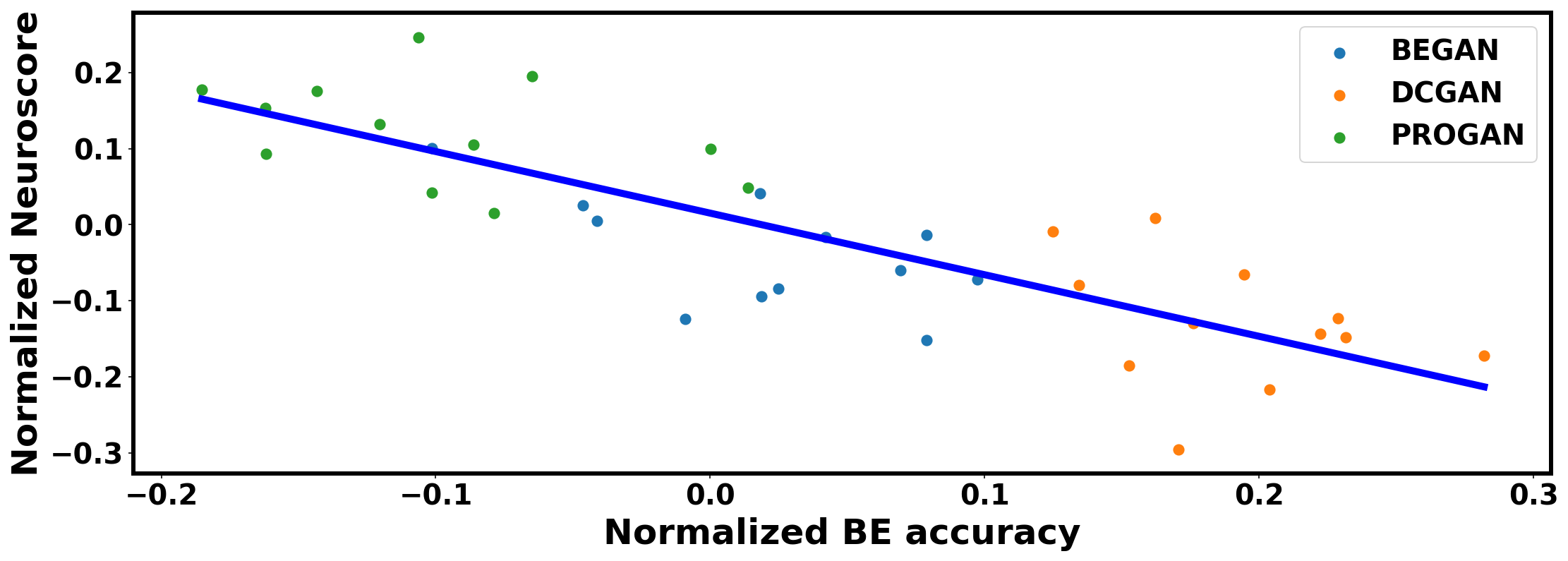}}
	\caption{Performance of real Neuroscore, calculated from participants' neural responses.} 
	\label{fig:Neuroscore_performance}
\end{figure}

The proposed Neuroscore reflects a human's perceptual response to different GANs via EEG measurements, thus it is consistent with human perceptual judgment on GANs. Figure~\ref{fig:Neuroscore_performance} demonstrates the performance of Neuroscore calculated from a human neural response. In Figure~\ref{fig:Neuroscore_performance}(a), it can be seen that different image categories activate different P300 responses. Figure~\ref{fig:Neuroscore_performance}(b) illustrates a strong correlation between Neuroscore and human judgment (BE accuracy)\footnote{BE accuracy is the recorded accuracy (calculated as the number of correctly labeled images divided by the total number of images) in the behavioral experiment. Normalized BE accuracy is calculated by subtracting the average accuracy (across GAN types for that participant) from BE accuracy.}. These results demonstrate that Neuroscore reflects human judgment perception. More details can be found in our previous work~\citep{wang2018use}.

\subsection{Neuro-AI Interface}
\begin{figure*}[!ht]
    \centering
    \includegraphics[width=1.\textwidth]{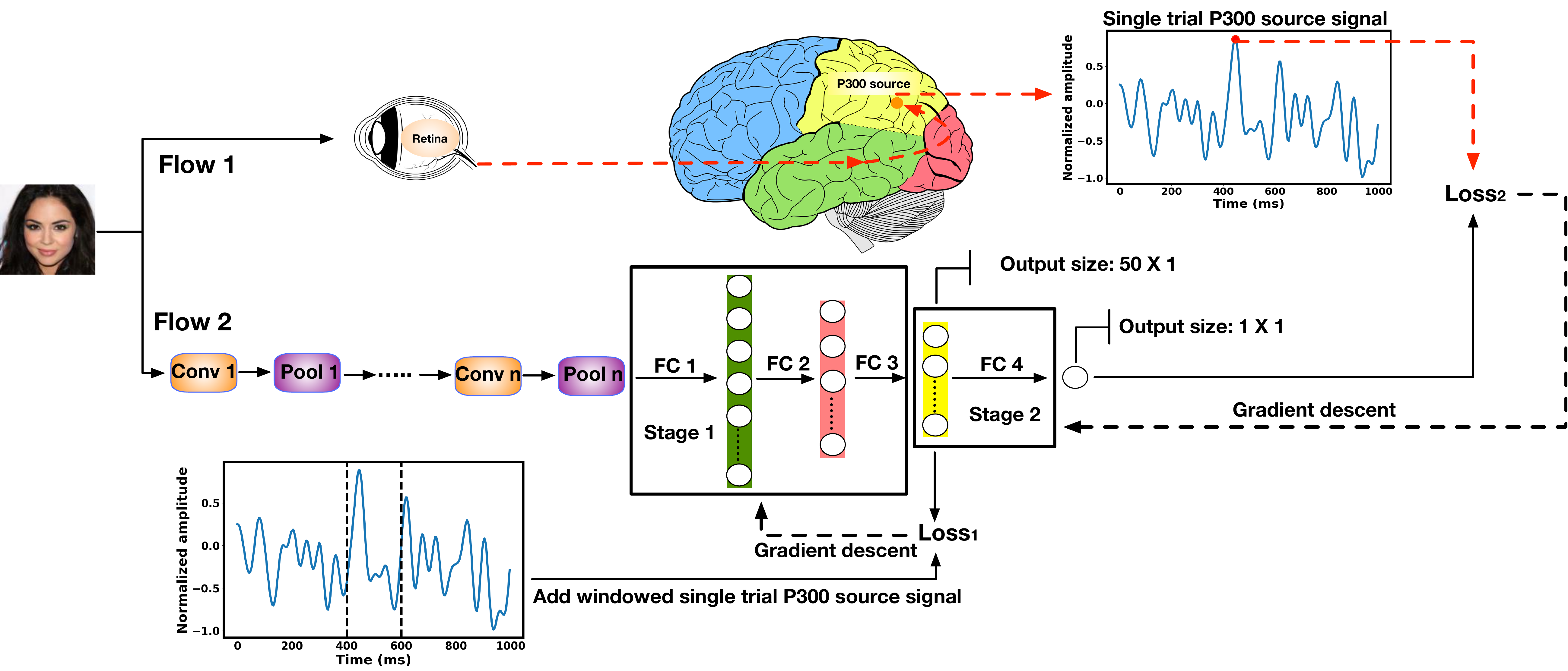}
    \caption{A neuro-AI interface and training details with added EEG information. Our training strategy includes two stages: (1) Learning from image to single-trial P300 source signal; and (2) Learning from single-trial P300 source signal to single-trial P300 amplitude. $\mathrm{loss}_{1}$ is the $\mathrm{L}_{2}$ distance between the yellow layer and the single-trial P300 source signal in the $400$ -- $600$ ms corresponding to the single input image. $\mathrm{loss}_{2}$ is the mean square error between model prediction and the single-trial P300 amplitude. $\mathrm{loss}_{1}$ and $\mathrm{loss}_{2}$ will be introduced in Section~\ref{sec:trianing-details}.}
    \label{fig:inspiration}
\end{figure*}

We propose a neuro-AI interface in order to generalize the use of Neuroscore. This kind of framework interfaces between neural responses and AI systems (a CNN is used in this study), which use neural responses as supervised information to train a CNN. The trained CNN is then used for generating a \textbf{synthetic-Neuroscore} given images generated by one of the popular GAN models i.e., average the outputs of corresponding images. Figure~\ref{fig:inspiration}. 
demonstrates the schematic of the neuro-AI interface used in this work\footnote{We understand that a human being's brain system is much more complex than demonstrated in this work and that information flow in the brain is not one-directional~\citep{she2016evaluating,she2018reduced}. Our framework can be further extended to be more biologically plausible.}. Flow 1 shows that the image processed by a human being's brain produces a single-trial P300 source signal for each input image. Flow 2 in Figure.~\ref{fig:inspiration} demonstrates a CNN with included EEG signals during the training stage. The convolutional and pooling layers process the image similarly as retina has done~\citep{mcintosh2016deep}. \textit{It should be noted that a CNN model is trained by using single images with their corresponding \textbf{single-trial} EEG information (including single-trial P300 signal and single-trial P300 amplitude\footnote{Single-trial P300 amplitude refers the maximum value in the 400 ms – 600 ms time window of a single-trial EEG signal. Details can be referred to our previous work~\citep{wang2018use}.). The averaged output of a trained model in terms of one image category can be represented as the \textbf{synthesized Neuroscore} (we refer to it as \textbf{synthetic-Neuroscore} in this paper)}}. Fully connected layers (FC) 1 -- 3 aim to emulate the brain's functionality that produces the EEG signal. The yellow dense layer in the architecture aims to predict the single-trial P300 source signal at $400$ -- $600$ ms in response to each image input. In order to help the model make a more accurate prediction for the single-trial P300 amplitude for the output, the single-trial P300 source signal at $400$ -- $600$ ms is fed to the yellow dense layer to learn parameters for the previous layers in the training step. The model was then trained to predict the single-trial P300 source amplitude (the red point shown in signal-trail P300 source signal of Figure~\ref{fig:inspiration}).

\subsection{Training Details} \label{sec:trianing-details}
\begin{figure}[!ht]
    \centering
    \includegraphics[width=\textwidth]{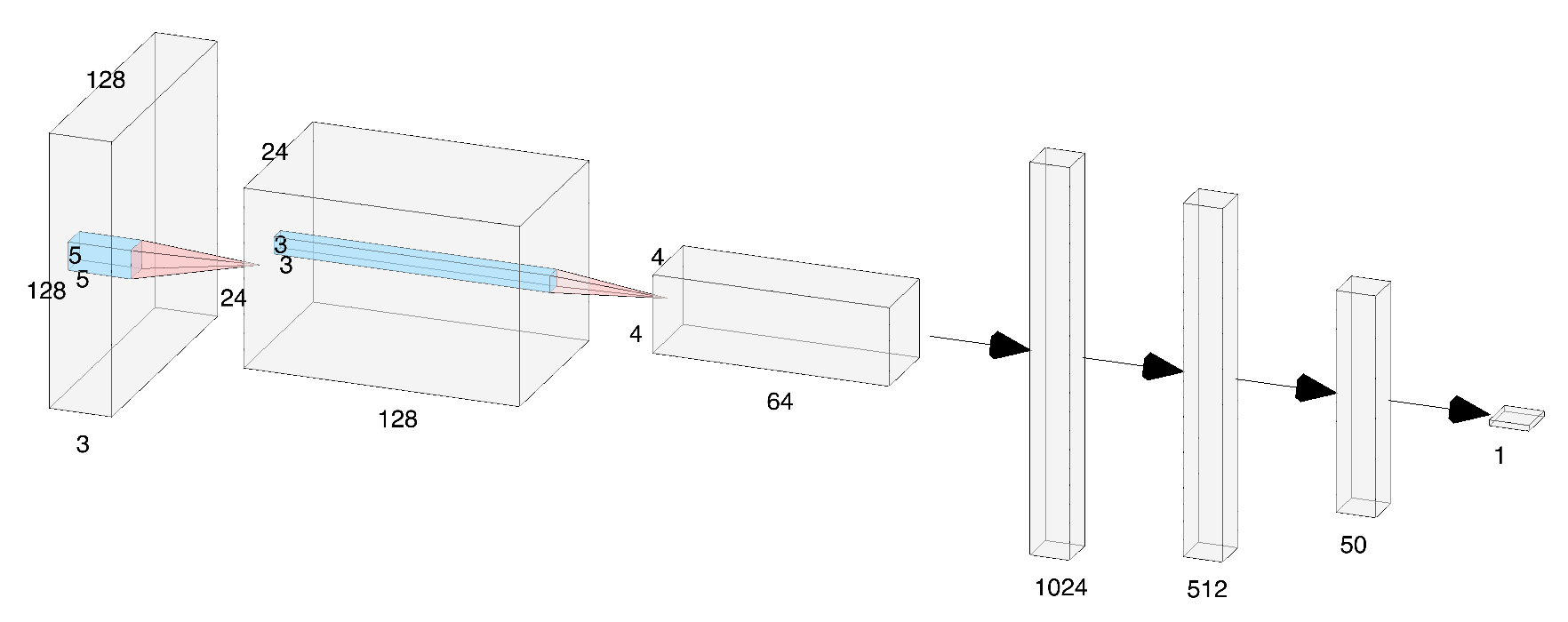}
    \caption{Shallow network architecture used in this work.}
    \label{fig:SCNN_architecture}
\end{figure}

Mobilenet V2, Inception V3 and Shallow network (architecture of Shallow network refers to Figure~\ref{fig:SCNN_architecture})
were explored in this work, where in flow 2 we use these three network bones such as Conv1-pooling layers. For Mobilenet V2 and Inception V3, we used ImageNet pre-trained parameters up to the FC 1 (as shown in Figure~\ref{fig:inspiration}). Table~\ref{tab:FC} shows the FC layers details of three networks. Due to no pretrained parameters in the Shallow net, only three FC layers are contained in order to avoid overfitting. We trained parameters from FC 1 to FC 4 for Mobilenet V2 and Inception V3. $\bm{{\mathrm{\theta}}}_{1}$ is used to denote the parameters from FC 1 to FC 3 and $\bm{{\mathrm{\theta}}}_{2}$ indicates the parameters in FC 4. For the Shallow model, parameters up to FC 2 represent $\bm{{\mathrm{\theta}}}_{1}$ and parameters in FC 3 indicate $\bm{{\mathrm{\theta}}}_{2}$.
\begin{table}[!htbp]
\centering
\begin{tabular}{c|c|c|c|c}
    \hline
    Model& FC 1& FC 2& FC 3& FC 4 \\ 
    Shallow net& (1024, 512)& (512, 50)& (50, 1)& NA\\
    Mobilenet & (1792, 896)& (896, 448)& (448, 50)& (50, 1)\\
    Inception & (2048, 1024)& (1024, 512)& (512, 50)& (50, 1)\\
    \hline
\end{tabular}
\caption{FC layers details of three networks investigated in this study.}
\label{tab:FC}
\end{table}

We added EEG to the model because we first want to find a function $f(\chi) \to \bm{\mathrm{s}}$ that maps the images space $\chi$ to the corresponding single-trial P300 source signal $\bm{\mathrm{s}}$. This prior knowledge can help us to predict the single-trial P300 amplitude in the second learning stage.

We compared the performance of the models with, without EEG signals and with randomized EEG signals for training. We defined two stage $\mathrm{loss}$ function ($\mathrm{loss}_{1}$ for a single-trial P300 source signal in the $400$ -- $600$ ms time window and $\mathrm{loss}_{2}$ for single-trial P300 amplitude) as
\begin{equation}
\begin{aligned}
 	\mathrm{loss_{1}}(\bm{\mathrm{\theta}}_{1}) &= \frac{1}{N} \sum_{i=1}^{N} \lVert \bm{\mathrm{S}}_{i}^{true} - \bm{\mathrm{S}}_{i}^{pred}(\bm{\mathrm{\theta}}_{1})\rVert ^{2}_{2}\\
 	\mathrm{loss_{2}}(\bm{\mathrm{\theta}}_{1},\bm{\mathrm{\theta}}_{2}) &= \frac{1}{N} \sum_{i=1}^{N}(\mathrm{y}_{i}^{true} - \mathrm{y}_{i}^{pred}(\bm{\mathrm{\theta}}_{1},\bm{\mathrm{\theta}_{2}}))^{2}
\end{aligned}
\end{equation}
where $\bm{\mathrm{S}}_{i}^{true} \in \mathbb{R}^{1 \times T}$ is the single-trial P300 signal in the $400$ - $600$ ms time window to the presented image, $\mathrm{y}_{i}$ refers to the single-trial P300 amplitude for each image, and $N$ refers to the batch size. In this case, we trained 20 epochs with batch size equaling to 256. An Adam optimizer with default hyperparameters was used and learning rate is 0.001.

The training of the models without using EEG signal is straightforward, models were trained directly to minimize $\mathrm{loss}_{2}(\bm{{\mathrm{\theta}}}_{1},\bm{{\mathrm{\theta}}}_{2})$ by feeding images and the corresponding single-trial P300 amplitude. In this case, training is an end-to-end process i.e., from an image to single-trial a P300 amplitude without considering stage 1. The reason that we do this is to investigate the significance of adding single-trial P300 signal as supervisory information to the network. Training with EEG information is explained in Algorithm~\ref{al:training}
\begin{algorithm}[!ht]
    \caption{Two training stages with EEG information.}\label{al:training}
    \begin{algorithmic}[1]
        \item[\textbf{Stage 1:} Training parameters $\bm{\mathrm{\theta}}_{1}$.]
        \Statex \textbf{Input:} Images and averaged P300 signal $\bm{\mathrm{S}}_{i}^{true}$.
        \For{number of training iterations}
            \State{Update $\bm{\mathrm{\theta}}_{1}$ by descending its stochastic gradient: $\nabla_{\bm{\mathrm{\theta}}_{1}} \frac{1}{N} \sum_{i=1}^{N} \lVert \bm{\mathrm{S}}_{i}^{true} - \bm{\mathrm{S}}_{i}^{pred}(\bm{\mathrm{\theta}}_{1})\rVert ^{2}_{2}$}
        \EndFor
        \item[\textbf{Stage 2:} Freezing $\bm{\mathrm{\theta}}_{1}$, training parameters $\bm{\mathrm{\theta}}_{2}$.]
        \Statex \textbf{Input:} Images and single-trial P300 amplitude $\mathrm{y}_{i}^{true}$.
        \For{number of training iterations}
            \State{Update $\bm{\mathrm{\theta}}_{2}$ by descending its stochastic gradient: $\nabla_{\bm{\mathrm{\theta}}_{2}} \frac{1}{N} \sum_{i=1}^{N}(\mathrm{y}_{i}^{true} - \mathrm{y}_{i}^{pred}(\bm{\mathrm{\theta}}_{1},\bm{\mathrm{\theta}}_{2}))^{2}$}
        \EndFor
    \end{algorithmic}
\end{algorithm}
and visualized in the ``Flow 2" of Figure~\ref{fig:inspiration} with two stages. Stage 1 learns parameters $\bm{{\mathrm{\theta}}}_{1}$ to predict P300 source signal while stage 2 learns parameters $\bm{\mathrm{\theta}}_{2}$ to predict single-trial P300 amplitude with $\bm{{\mathrm{\theta}}}_{1}$ fixed.

\section{Results}
\label{results}
\subsection{EEG and Model Performance} \label{sec:Models-performance}
\begin{figure}[!ht]
\centering
    \subfigure{
    \centering
    \includegraphics[width=.7\textwidth]{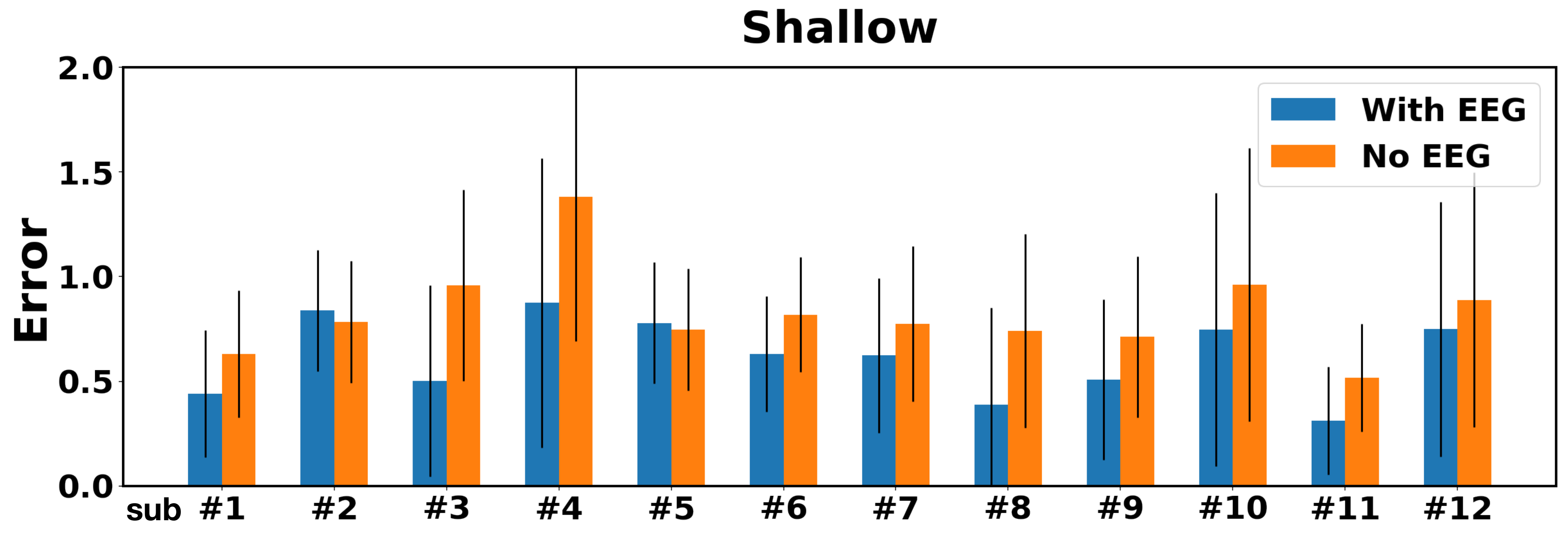}
    }
    \subfigure{
    \centering
    \includegraphics[width=.7\textwidth]{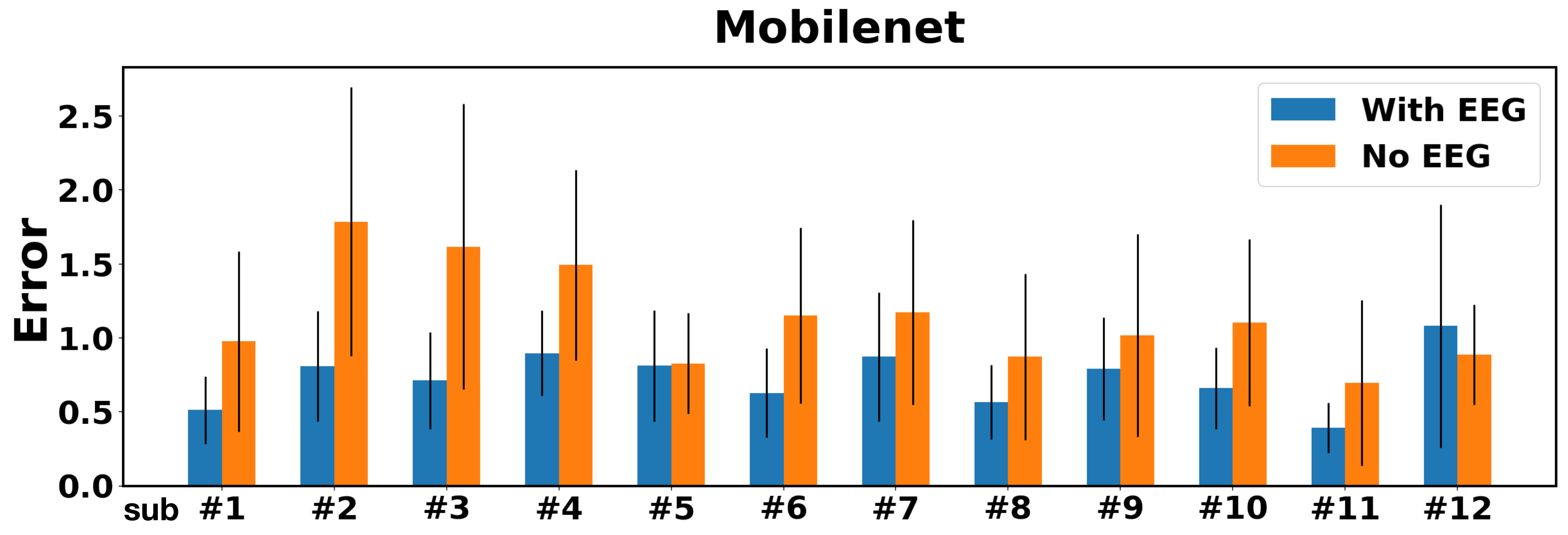}
    }
    \subfigure{
    \centering
    \includegraphics[width=.7\textwidth]{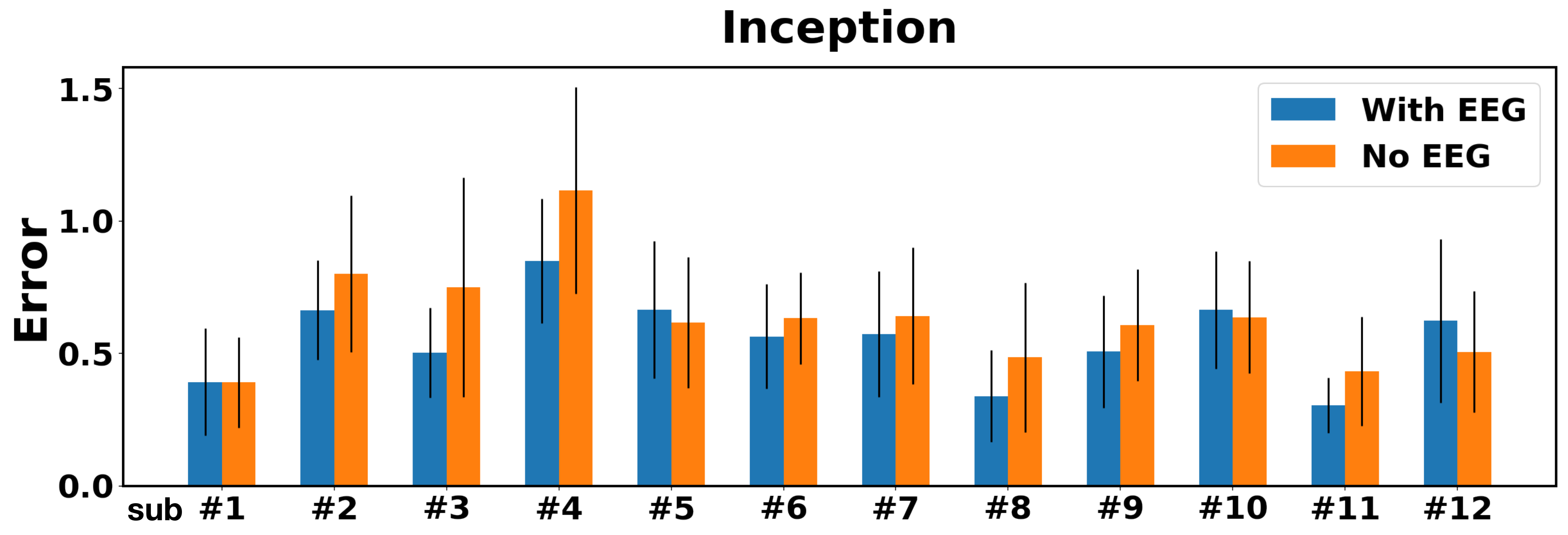}
    }
    \caption{Error of $3$ models with and without EEG signals. Error is defined as: $\sum_{i}^{m} \lvert \mathrm{Neuroscore}_{pred}^{(i)}  - \mathrm{Neuroscore}_{true}^{(i)} \rvert$, where $\mathrm{m} = 3$ is the number of GAN categories used (DCGAN, BEGAN, PROGAN, $12$ participants) and Neuroscore is obtained by averaging single-trial P300 amplitudes. A smaller value indicates better performance. Details of numeric values can be refered to Table\ref{tab:model_loss_individual}.}
    \label{fig:model_loss_individual}
\end{figure}

\paragraph{Individual Participant Performance}
Three models have been validated for each individual participant as shown in Figure~\ref{fig:model_loss_individual}. 
It can be seen that all three models trained with EEG outperform the models trained without EEG. \textbf{In other words, we show that including EEG/P300 time series signals as supervisory information to the yellow dense layer yields an improvement in performance as seen in Figure~\ref{fig:inspiration}}. with smaller error and standard deviation across almost all individual subjects. For those cases where the reverse is true (7 from 36 have better or equal performance without EEG), this might result from the number of EEG trials for an individual participant not being sufficient enough for training of deep networks to learn the mapping function $f(\chi)$ from image to EEG.
\begin{table}[!htbp]
\centering
\begin{tabular}{c|c|c}
\hline
\multicolumn{2}{c|}{Model} & Error mean (std) \\ \hline
\multirow{2}{*}{Shallow net} & Shallow-EEG & 0.151 ($\pm$0.245)\\ \cline{2-3} & Shallow & 0.428 ($\pm$0.623) \\ \hline
\multirow{2}{*}{Mobilenet} & Mobilenet-EEG & 0.155 ($\pm$0.235)\\ \cline{2-3} & Mobilenet & 0.437 ($\pm$0.589) \\ \hline
\multirow{2}{*}{Inception} & Inception-EEG & 0.157 ($\pm$0.487)\\ \cline{2-3} & Inception & 0.462 ($\pm$0.932) \\ \hline
\end{tabular}
\caption{Details of error mean and standard deviation for Figure~\ref{fig:model_loss_individual}.}
\label{tab:model_loss_individual}
\end{table}

\paragraph{Cross Participant Performance}
We evaluated the cross participant performance of our approach by pooling trials across participants to see if the use of pooled trials produced a smaller error. In this case, the number of EEG trials across participants is 6012. We split data into training and testing as 2:1 in which there are 4008 trials for training and 2004 trials for testing. All trials are randomly shuffled and we repeat this process for 20 times in order to get a more robust result.

Table~\ref{tab:model_loss_cross} shows the error for each model with the EEG signal, with a randomized EEG signal \textbf{within each type of GAN} and without an EEG signal. All models with EEG signals perform better than models without EEG signals, with much smaller errors and standard deviation.
\begin{table}[!htbp]
\centering
\begin{tabular}{c|c|c}
\hline
\multicolumn{2}{c|}{Model}                          & Error mean(std)    \\ \hline
\multirow{3}{*}{Shallow net} & Shallow-EEG           & \textbf{0.209 ($\pm$0.102)} \\ \cline{2-3} 
                             & Shallow-EEG$\mathrm{_{random}}$   & 0.348 ($\pm$0.114) \\ \cline{2-3} 
                             & Shallow               & 0.360 ($\pm$0.183) \\ \hline
\multirow{3}{*}{Mobilenet}   & Mobilenet-EEG         & \textbf{0.198 ($\pm$0.087)} \\ \cline{2-3} 
                             & Mobilenet-EEG$\mathrm{_{random}}$ & 0.404 ($\pm$0.162) \\ \cline{2-3} 
                             & Mobilenet             & 0.366 ($\pm$0.261) \\ \hline
\multirow{3}{*}{Inception}   & Inception-EEG         & \textbf{0.173 ($\pm$0.069)} \\ \cline{2-3} 
                             & Inception-EEG$\mathrm{_{random}}$ & 0.392 ($\pm$0.057) \\ \cline{2-3} 
                             & Inception             & 0.344 ($\pm$0.149) \\ \hline
\end{tabular}
\caption{Errors for $9$ models across the 12 participants (``*-EEG" indicates models are trained with paired EEG, ``*-EEG$\mathrm{_{random}}$" refers to EEG trials which are randomized in the $\mathrm{loss}_{1}$ \textbf{within each type of GAN}). Results are averaged by shuffling training/testing sets  $20$ times.}
	\label{tab:model_loss_cross}
\end{table}

Adding the EEG signal to the intermediate layer reduces error in all three models (as the same error is shown in Figure~\ref{fig:model_loss_individual}), namely $0.151$, $0.168$ and $\textbf{0.171}$ for Shallow, Mobilenet, and Inception respectively. This indicates that the Inception model benefits  most when adding EEG signal into the training stage. The performance of models with the EEG signal is ranked as Inception-EEG followwd by Mobilenet-EEG, and Shallow-EEG, which indicates that deeper neural networks may achieve better performance in this task. We used the randomized EEG signal here as a baseline to determine the efficacy of adding the EEG signal to produce better Neuroscore output. When randomizing the EEG signal, it shows that the error for each three model increases significantly. For Mobilenet and Inception, the error with the randomized EEG signal is even higher than those without the EEG signal in the training stage, demonstrating that EEG information in the training stage is crucial to each model.  

\begin{figure}[!ht]
	\centering
	\includegraphics[width=.7\textwidth]{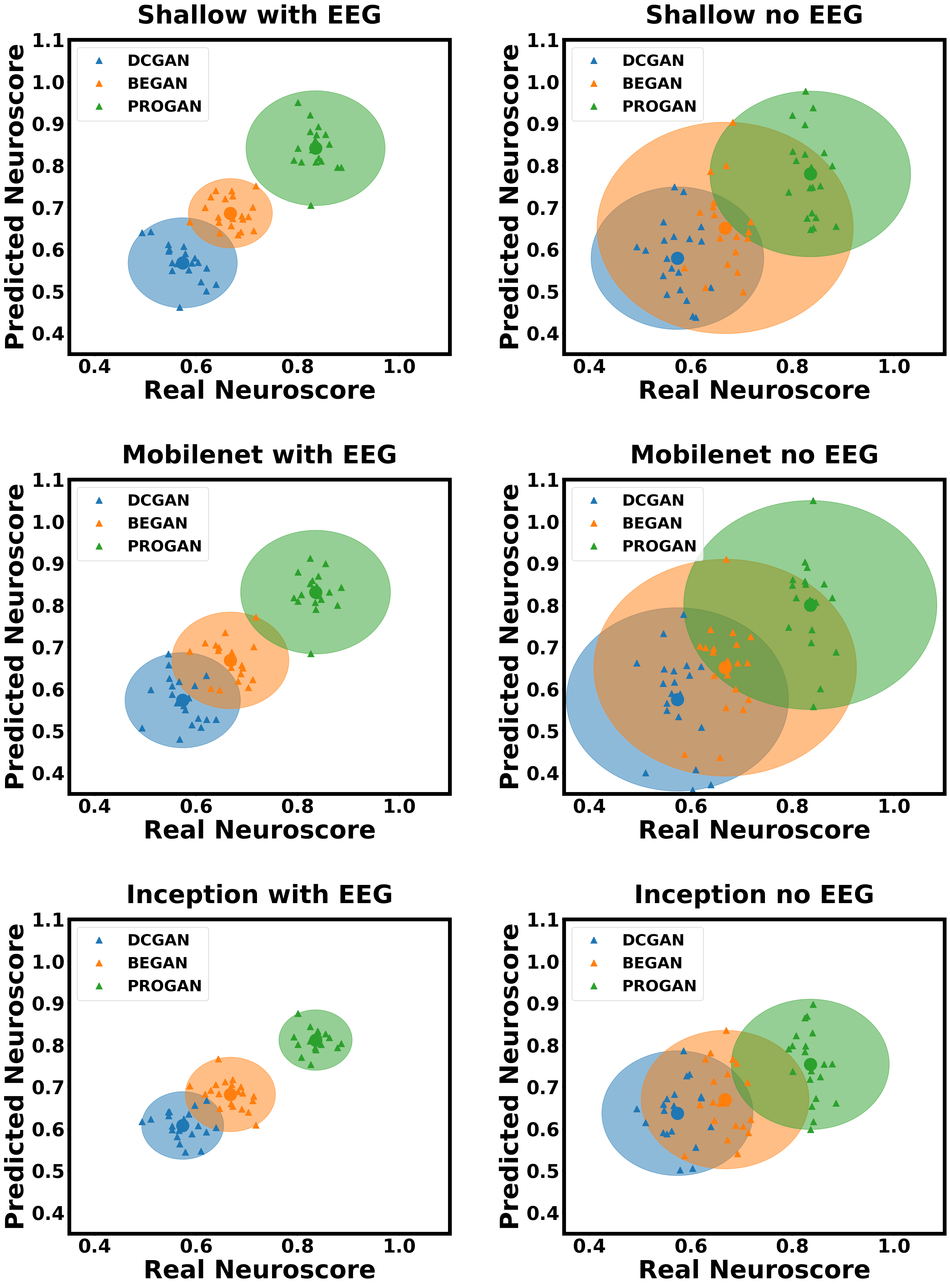}
	\caption{Scatter plot of synthetic-Neuroscore (vertical axis) and Neuroscore (horizontal) for $6$ models (Shallow, Mobilenet, Inception with and without EEG signals for training) across participants, with $20$ times repeated shuffling training and testing set. Each circle represents the cluster for a specific category. Small triangle markers inside each cluster correspond to each shuffling process. The dot at the center of each cluster is the mean.}  \label{fig:cross-subject-correlation}
\end{figure}

Figure~\ref{fig:cross-subject-correlation} shows that the models with EEG information have a stronger correlation between synthetic-Neuroscore and Neuroscore. The cluster (blue, orange, and green circles) for each category of the model trained with EEG (left column) is more separable than the cluster produced by model without EEG (right column). This indicates that when with EEG is used in training models Neuroscore is more accurate and that Neuroscore is able to rank the performances of different GANs, which cannot be achieved with other metrics~\citep{borji2018pros}.

\subsection{Neuroscore Aligns with Human Perception}
\label{sec:compare-to-other-metrics}
Figure~\ref{fig:Neuroscore_performance}(b) shows the correlation between Neuroscore and human judgment (BE accuracy) according to three GANs: BEGAN, DCGAN, and PROGAN. The statistical test demonstrates the strong correlation between those two variables. This indicates that Neuroscore can be used to evaluate GANs as it reflects human perceptual judgment. A number of previous studies have noted that increasing task difficulty reduces the amplitude of the P300 \citep{kim2008influence,marathe2013novel,senkowski2002effects,scharinger2017comparison}. It may be the case that the larger P300 amplitudes observed for the PROGAN images indicate that these face images were easier to detect compared to the images from the other GANs. For example, DCGAN images tended to contain far more visual aberrations and other inherent artefacts that would impede their detection \citep{wolfe2010reaction}. It has also been noted in another prior study that increased sensory evidence results in shorter reaction times and larger component amplitudes in temporal and spatial regions coinciding with those examined in our work \citep{philiastides2014human}. Another prior study explains larger P300 amplitudes for real face images resulting from enhanced perceptual processing \citep{schindler2017differential}. In effect, larger average P300/LPP amplitudes for a particular GAN type are indicative that its images are perceived as being real faces.

We have already demonstrated that the Neuroscore derived from raw EEG is consistent with human perception~\citep{wang2018use}. We will now demonstrate the same property of synthetic-Neuroscore predicted from the neuro-AI interface. We compare the synthetic-Neuroscore with three widely used evaluation metrics. The ultimate goal of GANs is to generate images that are indistinguishable from real images by human beings. Therefore, consistency between an evaluation metric and human perception is a critical requirement for the metric to be considered good. Table~\ref{tab:traditional-method-score} shows the comparison between synthetic-Neuroscore and three traditional scores. To be consistent with all the scores (smaller score indicates better GAN), we used 1/IS and 1/synthetic-Neuroscore for comparisons in Table~\ref{tab:traditional-method-score}. It can be seen that people rank the GAN performance as PROGAN $>$ BEGAN $>$ DCGAN. All three synthetic-Neuroscores produced by the three models with EEG are consistent with human judgment while the other three conventional scores are not (they all indicate that DCGAN outperforms BEGAN).

\begin{table}[ht!]
\small 
\centering
\begin{tabular}{c|c|c|c|c}
\hline
\multicolumn{2}{c|}{Metrics}           & DCGAN & BEGAN & PROGAN \\ \hline
\multicolumn{2}{c|}{1/IS}              &  {0.44} & {0.57} & {0.42}    \\ \hline
\multicolumn{2}{c|}{MMD}               &  {0.22} & {0.29}  & {0.12}      \\ \hline
\multicolumn{2}{c|}{FID}               &  {63.29} & {83.38} & {34.10}       \\ \hline
\multirow{3}{*}{\textbf{Proposed Methods}} & 1/Shallow-EEG   &  {\textbf{1.60}} & {\textbf{1.39}} & {\textbf{1.14}}    \\ \cline{2-5} 
                      & 1/Mobilenet-EEG & {\textbf{1.71}} & {\textbf{1.29}} & {\textbf{1.20}}   \\ \cline{2-5} 
                      & 1/Inception-EEG & {\textbf{1.51}} & {\textbf{1.34}} & {\textbf{1.24}}  \\ \hline
\multicolumn{2}{c|}{Human (BE accuracy)} &{\textbf{0.995}} & {\textbf{0.824}} & {\textbf{0.705}}     \\ \hline
\end{tabular}
\caption{Three conventional scores: Inception Score (IS), Maximum Mean Discrepancy (MMD), Fr\'{e}chet Inception Distance (FID), and synthetic-Neuroscore produced by three models with EEG for each GAN category. A lower score indicates better performance of the GAN. Neuroscore is consistent with human judgments. Bold text indicates the consistency with human judgment (BE) accuracy.}
	\label{tab:traditional-method-score}
\end{table}

\subsection{Synthetic-Neuroscore Needs Far Fewer Samples}
\begin{figure}[!ht]
    \centering
    \includegraphics[width=.6\textwidth]{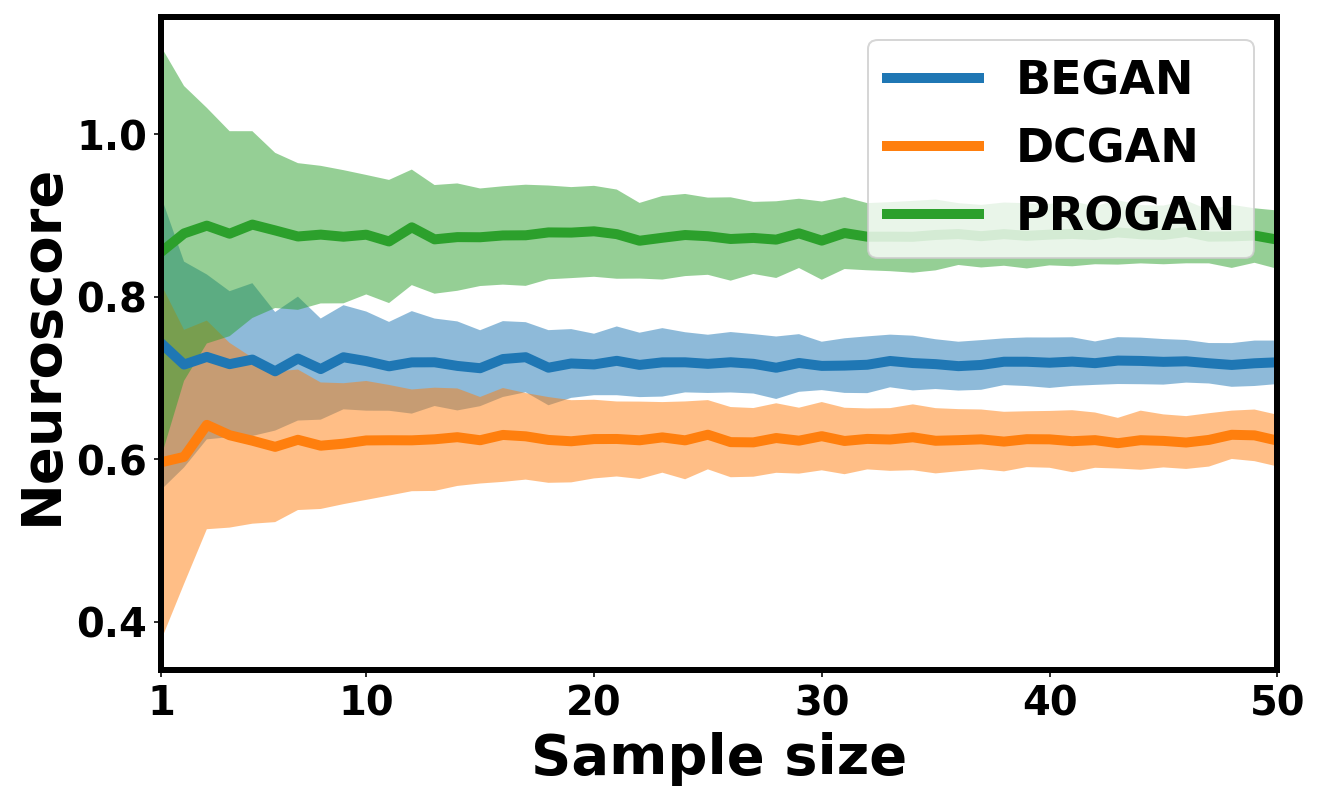}
    \caption{Synthetic-Neuroscore for different evaluated sample sizes for each type of GAN. 200 repeated measurements have been made by randomly shuffling image samples.}\label{fig:gen-test}
\end{figure}

The number of samples needed for evaluation of a GAN is crucial in real-world applications considering computational efficiency and efforts needed for labeling and annotation. Traditional metrics need a large sample size to capture the underlying statistical properties of the real and generated images~\citep{salimans2016improved,empirical-study}. In practice, we should prefer a metric that is robust when dealing with small sample sizes i.e., where small sample sizes can produce good estimates. Figure~\ref{fig:gen-test}(b) shows that synthetic-Neuroscore converges stably at around $20$ presentations of a specific image (for signal-enhancement purposes), which is far fewer than the thousands of images required by traditional methods~\citep{borji2018pros,empirical-study}. This is due to the fact that the LDA-beamformed single-trial P300 amplitude becomes stable when as few as dozens of EEG trials corresponding to one category are available~\citep{mouraux2008across}.


\subsection{Synthetic-Neuroscore Can Rank Images}
\begin{figure}[!ht]
    \centering
    \includegraphics[width=.7\textwidth]{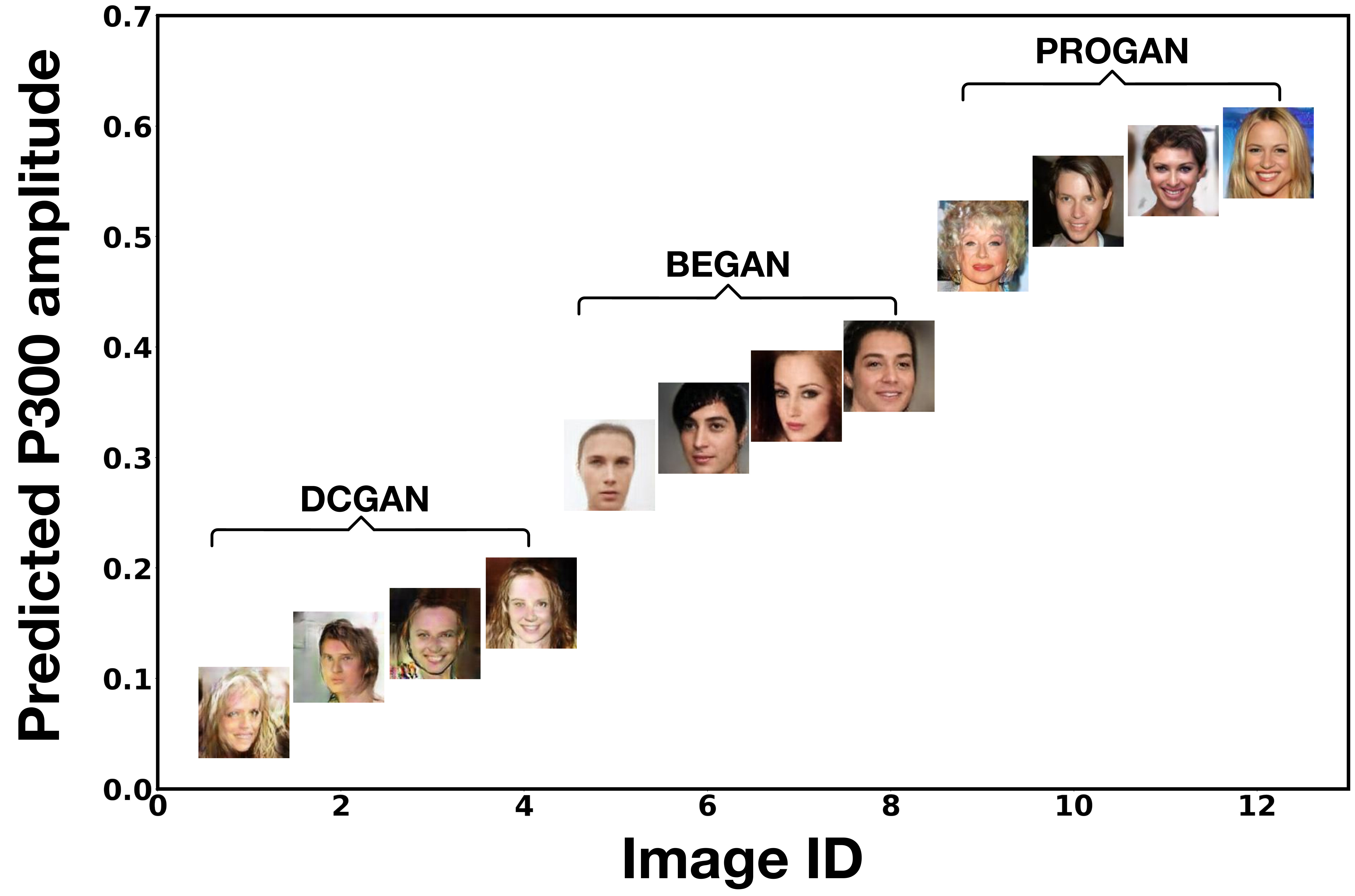}
    \caption{P300 for each single image predicted by the proposed neuro-AI interface in our paper. Higher predicted P300 indicates  better image quality.}\label{fig:rank-single-img}
\end{figure}

Another property of using synthetic-Neuroscore is the ability to indicate the quality of an individual image. Traditional evaluation metrics are unable to score each individual image for two reasons. Firstly they need large-scale samples for evaluation and secondly most methods (e.g., MMD and FID) evaluate GANs based on the dissimilarity between real images and generated images so they are not able to score the generated images individually. For our proposed method, the score of each single image can also be evaluated as a synthetic single-trial P300 amplitude measurement. We demonstrate in Figure~\ref{fig:rank-single-img} how the predicted single-trial P300 amplitude conveys perceptual quality at the level of individual images.
This property provides synthetic-Neuroscore with a novel capability for tracking variations in image output quality within a typical GAN. Although synthetic-Neuroscore and IS are both generated from deep neural networks, synthetic-Neuroscore is more suitable than IS for evaluating GANs as it is a direct reflection of human perception and  fewer sample images are required for evaluation.  This has benefits in terms of improved  explanation of output than that offered by IS.  For example low ranked images can be selected at evaluation time to illustrate cases where the GAN under evaluation is performing poorly.

%
%
%


\section{Conclusions}
In this paper, we outline a metric for evaluating the quality of the outputs from GANs called Neuroscore. Furthermore, we describe a neuro-AI interface to calculate a synthetic-Neuroscore for evaluating GAN performance that only requires EEG signals as supervisory information during model training. Three deep network architectures are explored and our results demonstrate that including neural responses during the training phase of the neuro-AI interface improves its accuracy even though neural measurements are absent when evaluating on a test set. This means that human subjects are not actually needed to evaluate the output from a test GAN, their neural responses are needed only when training the model that produces a synthetic-Neuroscore.

We compared our synthetic-Neuroscore measure to three traditional evaluation metrics and demonstrated the unique advantages of synthetic-Neuroscore, that it is consistent with human perception, that it requires far fewer image samples for calculation and that it can rank individual images in terms of quality, within a specific GAN.

In this work, we demonstrated the use of CNNs to synthesize the neural response. More complicated neural architectures such as mixture of CNNs and recurrent neural networks can be investigated in  future work when more EEG data is available.

\section*{Acknowledgements}
This work is funded as part of the Insight Centre for Data Analytics which is supported by Science Foundation Ireland under Grant Number SFI/12/RC/2289.








\clearpage
\bibliographystyle{elsarticle-num}
\bibliography{ref}

\end{document}